\documentclass{article} 
\usepackage{iclr2027_conference,times}

\usepackage{amsmath,amsfonts,bm}

\def\eqref#1{equation~\ref{#1}}

\def\1{\bm{1}}

\DeclareMathAlphabet{\mathsfit}{\encodingdefault}{\sfdefault}{m}{sl}
\SetMathAlphabet{\mathsfit}{bold}{\encodingdefault}{\sfdefault}{bx}{n}

\newcommand{\Cov}{\mathrm{Cov}}

\usepackage{hyperref}
\usepackage{url}
\usepackage{booktabs}
\usepackage{graphicx}
\usepackage{multirow}
\usepackage{wrapfig}
\usepackage{arydshln}
\usepackage{paralist}
\usepackage{subcaption}
\usepackage{fvextra}
\usepackage{tcolorbox}
\tcbuselibrary{breakable}
\usepackage{enumitem}
\usepackage{amsmath}
\usepackage{amssymb}
\usepackage{amsthm}
\usepackage{xurl}

\AtBeginDocument{%
  \setlength{\abovedisplayskip}{3pt plus 1pt minus 2pt}%
  \setlength{\belowdisplayskip}{3pt plus 1pt minus 2pt}%
  \setlength{\abovedisplayshortskip}{0pt plus 1pt}%
  \setlength{\belowdisplayshortskip}{2pt plus 1pt minus 1pt}%
  \setlength{\jot}{2pt}%
}
\renewcommand{\eqref}[1]{Eq.~(\ref{#1})}

\newcommand{\ours}{\textsc{GRPODropout}}
\title{\ours: Less is More for Online Reinforcement Learning Rollouts}

\author{
  Hexuan Deng\textsuperscript{\rm 1, 2}$^\dag$~~~
  Zihao Yan\textsuperscript{\rm 1}$^\dag$~~~
  ~Xuebo Liu\textsuperscript{\rm 1}\thanks{~~Xuebo Liu is the corresponding author.}~~~
  Shuo Nie\textsuperscript{\rm 1}~~~
  Yue Wang\textsuperscript{\rm 2,3}~~~
  \textbf{Chen Wang}\textsuperscript{\rm 2}
  \\
  ~\textbf{Zhaohua Zhang}~~~
  \textbf{Tianwen Jiang}~~~
  \textbf{Qiuyong Xiao}~~~
  \textbf{Jihong Zhang}~~~
  \textbf{Min Zhang}\textsuperscript{\rm 1}
  \\
  \textsuperscript{\rm 1} Institute of Computing and Intelligence, Harbin Institute of Technology (Shenzhen)
  \\
  \textsuperscript{\rm 2} Beijing Zhongguancun Academy
  \quad \textsuperscript{\rm 3} XinzhuAI
}

\newcommand{\old}{\mathrm{old}}

\newtheorem{proposition}{Proposition}

\iclrfinalcopy 
\begin{document}

\maketitle

\begin{abstract}
Reinforcement learning (RL) methods such as GRPO substantially improve large language model reasoning but often suffer from policy entropy collapse: the loss of sampling diversity weakens exploration and limits further improvement. Existing methods address this issue either through algorithm-level interventions, such as reward modification and entropy/KL regularization, or through token-level reweighting. We investigate a complementary perspective: entropy collapse can also be mitigated by changing which generated rollouts contribute to policy updates. Under the same sampling budget, not all rollouts contribute positively to an update, and selectively excluding some can improve learning. To address this, we propose \ours: before the standard update, we use a simple strategy that selectively removes a small number of high-probability positive-advantage rollouts and recenters the retained advantages. To motivate this design, we develop a rollout-level theoretical analysis that guides method design and threshold selection. The method changes only rollout usage, and adds negligible computational overhead. Experiments show higher accuracy than original GRPO and higher actor entropy while using fewer rollout samples for updates, illustrating ``less is more.'' This work provides insight into RL rollout usage: removing some rollouts can improve performance. Code is available at~\url{https://github.com/hexuandeng/GRPODropout/}.
\end{abstract}

\section{Introduction}

Online RL substantially improves the reasoning ability of large language models~\citep{shao2024deepseekmath,yu2025dapo}. However, these methods often suffer from policy entropy collapse during training, reducing sampling diversity and weakening exploration of alternative reasoning paths~\citep{cui2025entropymechanism,wang2025revisitingentropy}. Because online RL relies on newly sampled rollouts for subsequent updates, this reduced diversity can limit opportunities to discover new solutions and improve further~\citep{yue2025reasoningcapacity}. This highlights a fundamental challenge in RL: the exploration--exploitation trade-off~\citep{auer2002finite}.

To address this problem, existing approaches follow two routes. The first modifies the RL algorithm through rewards or advantages~\citep{hao2025steer,wang2025revisitingentropy}, clipping bounds~\citep{yu2025dapo,cliplowhigh2025}, or temperature and sampling control~\citep{wang2025aepo}. The second focuses on data usage at the token level, targeting high-entropy tokens and critical branching points for masking or reweighting~\citep{du2025ulorl,cui2025entropymechanism,wang2025beyond8020}, branching~\citep{li2025cure,rose2026}, or calibration~\citep{yang2026paec}. These methods exploit differences in token importance to mitigate entropy collapse. Here, we argue that changing rollout usage alone can mitigate entropy collapse. However, existing work still offers limited theoretical and methodological guidance for selecting complete rollouts to balance exploration and exploitation.

To bridge this gap, we propose \ours{}, which selectively removes rollouts to improve performance, mitigate entropy collapse, and preserve exploration in the trained model. Specifically, under our rollout-level theoretical analysis, we find that high-probability positive rollouts and low-probability negative rollouts can both reduce policy entropy, with the former having a more pronounced effect. Intuitively, high-probability positive rollouts are already well represented under the current policy; further reinforcement may provide limited additional information while accelerating entropy collapse. We therefore selectively remove these rollouts and recenter the retained advantages using old-policy probability weights to correct the imbalance introduced by deletion. Across three models and ten benchmarks, the average score improves over original GRPO by 1.55 percentage points. Despite using 12.1\% fewer retained rollouts in policy updates on average in groups with nonzero advantages, training and held-out actor entropy are 446.1\% and 256.0\% higher than original GRPO, respectively, at the end of training, demonstrating better results with fewer update rollouts. Our contributions are:
\begin{itemize}[leftmargin=1.2em,itemsep=0pt,topsep=0pt]
\item \textbf{A rollout-level analysis of data importance.} We extend the theoretical analysis to the rollout level to motivate our approach, and use experiments to further characterize which rollouts can reduce policy entropy, providing insight into how rollouts are used during RL training.
\item \textbf{Rollout filtering and recentering.} We design a data-selection algorithm to filter potentially harmful rollouts and recenter retained advantages to correct the imbalance introduced by deletion.
\item \textbf{Empirical evidence across model sizes.} We evaluate accuracy and entropy across three model sizes under reduced update-level rollout usage.
\end{itemize}

\section{Related Work}

\paragraph{Token-level entropy interventions.}
One line of work stabilizes policy entropy by identifying less informative tokens and attenuating their updates. UloRL~\citep{du2025ulorl} identifies high-probability tokens in successful rollouts as already well learned and masks them when policy entropy falls below a target. RSI-S~\citep{lv2026rsi} extends this idea by normalizing sampled-token surprisal by local entropy. \citet{cui2025entropymechanism} analyze first-order entropy changes and use this analysis to motivate Clip-Cov and KL-Cov, which clip or penalize updates on high-covariance tokens. \citet{wang2025beyond8020} show that updating only the 20\% highest-entropy tokens can match or outperform full-token training. Later methods identify critical branching points to construct more diverse training data. CURE~\citep{li2025cure} and ROSE~\citep{rose2026} branch at positions where likely next tokens are semantically distinct and train on the resulting trees using segment-level advantages. ERPO~\citep{yu2026erpo} detects Critical Decision Pivots using group-normalized entropy and combines entropy gating, reasoning progress, and final outcomes into token-level advantages. PAEC~\citep{yang2026paec} combines normalized entropy with top-two competition to construct a detached soft position mask. STARE~\citep{luo2026stare} derives an advantage--surprisal four-quadrant structure from first-order entropy variation and uses target-entropy feedback to regulate high-surprisal tokens. ICT~\citep{feng2026ict} selects positions whose full next-token distributions exhibit large Jensen--Shannon divergences, capturing both high- and low-entropy forks.

These token-level interventions leave open how entropy contributions can guide the selective removal of complete rollouts sampled for each prompt. We address this gap through rollout-level entropy analysis and principled rollout selection, as unmasked tokens may still reinforce existing paths and limit exploration~\citep{du2025ulorl,zhu2025negative}.

\paragraph{Entropy-aware reinforcement learning algorithms.}
Another line addresses entropy collapse by modifying the training signal or optimization rule. One group modifies rewards or advantages. STEER~\citep{hao2025steer} estimates the first-order entropy change induced by each token update and downweights updates that sharply decrease or increase entropy. \citet{wang2025revisitingentropy} identify positive-advantage tokens as the main source of entropy reduction and reweight them according to the training stage or current entropy. HAPO~\citep{he2026eapo} reallocates advantages toward high-entropy positions through capacity-guided credit assignment. Papo~\citep{papopaper2026} predicts the update direction from the advantage sign, sampled-token probability, and local distribution, and regulates opposing entropy polarities by training phase.
A second group modifies the optimization rule directly. DAPO~\citep{yu2025dapo} introduces Clip-Higher to relax the upper importance-ratio bound. Clip-Low/High~\citep{cliplowhigh2025} show the underlying optimizer bias: lower clipping tends to increase entropy, whereas upper clipping decreases it. SCOPE-RL~\citep{wang2025aepo} uses temperature-adaptive positive samples and an auxiliary regularizer to control policy entropy.

These methods mitigate entropy collapse through algorithmic changes to rewards, advantages, clipping, or regularization. We find that data-level interventions alone can also mitigate entropy collapse, through rollout selection and advantage recentering while preserving the RL objective form.

\section{Method}

\begin{figure}[t]
\centering
\includegraphics[width=0.9\linewidth]{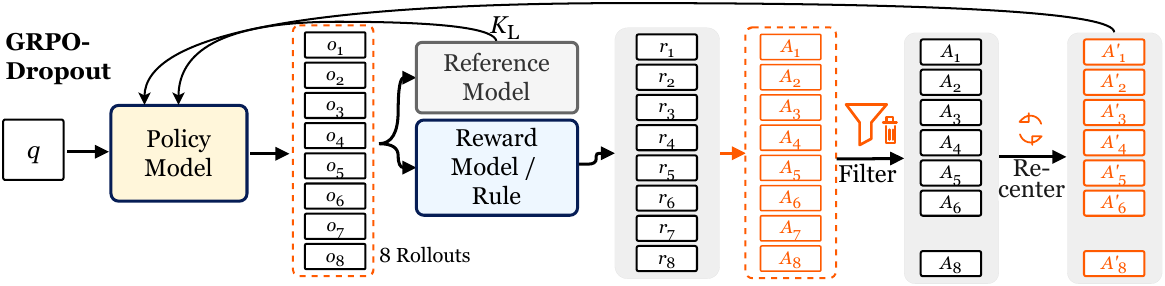}
\caption{GRPODropout procedure. Filter removes only selected positive-advantage rollouts; Re-center applies probability-weighted advantage recentering. Other GRPO updates are unchanged.}
\label{fig:method-overview}
\end{figure}

We find that not all rollouts benefit model training and demonstrate that removing a subset of the data can improve performance. We therefore propose \ours{}, as illustrated in Figure~\ref{fig:method-overview}. Each subsection presents a local theoretical analysis to motivate the corresponding design choice, while experiments evaluate the complete method. Section~\ref{sec:cov-deletion} develops rollout-level first-order entropy analysis and candidate deletion; Section~\ref{sec:positive-only} motivates positive-only deletion through a conditional comparison; Section~\ref{sec:renorm} analyzes deletion-induced drift of zero-advantage sequences, derives recentering, and presents the full procedure.

\suppressfloats[t]

\subsection{Preliminaries}
\label{sec:setup}

\paragraph{Problem Formulation.}
A rollout is a complete output sequence generated by the policy for a given prompt. For a fixed prompt $x$, let $\pi_{\old}$ denote the policy before the policy update, and let $\mathcal Y_x=\{\mathbf y_i\}_i$ be the set of all possible complete output sequences generated from $x$ under $\pi_{\old}$. For this fixed prompt, its Shannon entropy is $H(\pi_{\old})=-\mathop{\smash[b]{\sum\nolimits_{\mathbf y_i\in\mathcal Y_x}}}\nolimits\pi_{\old}(\mathbf y_i\mid x)\log\pi_{\old}(\mathbf y_i\mid x)$. We seek to optimize task return while mitigating policy entropy collapse and preserving exploration. For policy updates, we independently sample a training group of $G$ complete rollouts from this space according to the old policy: $\smash[t]{\mathbf y^{(1)}},\ldots,\smash[t]{\mathbf y^{(G)}}\mathrel{\smash[t]{\overset{\mathrm{i.i.d.}}{\sim}}}\pi_{\old}(\cdot\mid x)$. Here $i$ indexes the space of complete output sequences, whereas $g\in\{1,\ldots,G\}$ indexes the sampled group. Rollout $g$ is $\smash[t]{\mathbf y^{(g)}}=(\smash[t]{y_1^{(g)}},\ldots,\smash{y_{T_g}^{(g)}})$, with length $T_g$, giving sequence probability $\pi_{\old}(\mathbf y^{(g)}\mid x)=\prod_{t=1}^{T_g}\pi_{\old}(y_t^{(g)}\mid x,\mathbf y_{<t}^{(g)})$.

\paragraph{Policy optimization.}
Let $\pi_\theta$ be the policy to optimize, with parameters $\theta$. Given a rollout reward $R(x,\mathbf y)$, let $A(x,\mathbf y)$ denote advantage relative to a baseline $b(x) = \bar{R}(x, *)$, e.g., $R(x,\mathbf y)-b(x)$. Using the old policy for sampling, we write the advantage objective with reference-policy regularization~\citep{stiennon2020summarize,ouyang2022instructgpt} and its sampled surrogate as
\begin{equation}
\begin{aligned}
&\max_\theta\Bigl\{\mathbb E_{\mathbf y\sim\pi_{\old}(\cdot\mid x)}\!\Bigl[
\frac{\pi_\theta(\mathbf y\mid x)}{\pi_{\old}(\mathbf y\mid x)}A(x,\mathbf y)\Bigr]
-\beta\,\mathrm{KL}\!\bigl(\pi_\theta(\cdot\mid x)\,\|\,\pi_{\mathrm{ref}}(\cdot\mid x)\bigr)\Bigr\}\\
&\leadsto\quad\max_\theta\Bigl\{\frac1G\sum_{g=1}^G A^{(g)}
\frac{\pi_\theta(\mathbf y^{(g)}\mid x)}{\pi_{\old}(\mathbf y^{(g)}\mid x)}
-\beta\,\mathrm{KL}\!\bigl(\pi_\theta(\cdot\mid x)\,\|\,\pi_{\mathrm{ref}}(\cdot\mid x)\bigr)\Bigr\}.
\end{aligned}
\label{eq:advantage-surrogate}
\end{equation}
Here $\pi_{\mathrm{ref}}$ is a fixed reference policy, $\beta\ge0$ is its regularization coefficient, and $\mathrm{KL}$ denotes Kullback--Leibler divergence. The probability-ratio identity assumes that the candidate policy is supported within the old policy and the advantage expectation exists. The arrow denotes Monte Carlo approximation using sampled group of rollouts and advantages $A^{(g)}$.

\subsection{Rollout-Level Entropy Effect and Candidate Deletion}
\label{sec:cov-deletion}

In this subsection, we analyze which rollouts decrease policy entropy under a local update model. Motivated by this analysis and intuition, we design a filtering rule to mitigate entropy collapse.

\paragraph{Theoretical motivation: rollout updates and entropy collapse.}
Reward optimization can improve return while increasing the risk of entropy collapse and loss of exploration. To identify which rollouts decrease entropy, we first characterize each rollout's first-order entropy contribution and then give the criterion for the overall entropy change.

To extend token-level entropy analysis to complete rollouts, we analyze a local KL-proximal policy update following~\citet{abdolmaleki2018maximum}. With the sampled group and its advantages fixed, let $\pi^{(\eta)}$ be the updated policy at step size $\eta$, with $\pi^{(0)}=\pi_{\old}$. Under this model, we obtain the following results:

\begin{proposition}[First-order entropy contribution of a single rollout]
\label{prop:single-rollout}
Fix a prompt $x$ and adopt the local update model and assumptions in Appendix~\ref{app:proofs}. Let $s_g=-\log\pi_{\old}(\mathbf y^{(g)}\mid x)$ be the surprisal of sampled rollout $g$ under the old policy. Its additive first-order entropy contribution to the whole-group update is
\begin{equation}
h_g=\frac{A^{(g)}}G\bigl(s_g-H(\pi_{\old})\bigr).
\label{eq:single-contribution}
\end{equation}
\end{proposition}

\begin{proposition}[Entropy derivative from sampled rollouts]
\label{prop:entropy-covariance}
Under the setting of Proposition~\ref{prop:single-rollout}, assume $\sum_{g=1}^G A^{(g)}=0$. Using the surprisals $s_g$ defined above, let $\bar s=G^{-1}\mathop{\smash{\sum\nolimits_{g=1}^G}}\nolimits s_g$. For the fixed sampled group, the proximal update has entropy derivative
\begin{equation}
\left.\frac{dH(\pi^{(\eta)})}{d\eta}\right|_{\eta=0}
=\operatorname{Cov}_{g=1}^{G}\!\bigl(A^{(g)},s_g\bigr)
=\frac1G\sum_{g=1}^G A^{(g)}\bigl(s_g-\bar s\bigr).
\label{eq:covariance-conclusion}
\end{equation}
\end{proposition}

The proofs are in Appendix~\ref{app:proofs}. Proposition~\ref{prop:single-rollout} shows that a positive-advantage rollout with $s_g<H(\pi_{\old})$ (relatively high probability) and a negative-advantage rollout with $s_g>H(\pi_{\old})$ (relatively low probability) both have negative first-order entropy contributions. Conversely, low-probability positives and high-probability negatives contribute to entropy increase. Intuitively, further reinforcing already likely positive-advantage rollouts or suppressing already unlikely negative-advantage rollouts may offer limited gains in reasoning ability. Repeatedly prioritizing such updates may instead accelerate entropy collapse and weaken exploration.

Proposition~\ref{prop:entropy-covariance} shows that the proximal update decreases entropy to first order when rollout advantages and log-probabilities are positively correlated, which is often indeed tend to be positively correlated in original models~\citep{kadavath2022language,zenn2026likely}, providing an intuitive explanation for entropy collapse during reward optimization. Further, during RL training, by further increasing the likelihood of positive-advantage rollouts and decreasing that of negative-advantage ones, reward optimization can further reinforce this positive association and intensify entropy reduction over training.

\paragraph{Method: candidate filtering.}
Based on the preceding analysis, we design a rollout-selection method that uses Proposition~\ref{prop:single-rollout} to identify candidates with negative estimated entropy contributions and Proposition~\ref{prop:entropy-covariance} to set the deletion threshold. Specifically, we reuse $h_g$ from \eqref{eq:single-contribution}, estimating $H(\pi_{\old})$ by the Monte Carlo mean $\bar s$ of the original sampled group. In the filtering rule below, $h_g$ denotes the contribution computed with this substitution. Using Proposition~\ref{prop:entropy-covariance}, we employ a greedy algorithm to select the largest retained set that ensures the estimated entropy does not decrease:
\begin{equation}
\mathcal I_{\mathrm{keep}}\in\arg\max\nolimits_{\mathcal I\subseteq\{1,\ldots,G\}}|\mathcal I|
\quad\text{s.t.}\quad
\operatorname{Cov}_{g=1}^{G}\!\bigl(A^{(g)}\mathbf 1\{g\in\mathcal I\},s_g\bigr)\ge0.
\label{eq:cov-contribution-selection}
\end{equation}
Here $\mathcal I_{\mathrm{keep}}$ is the set of rollout indices selected for retention. The covariance is evaluated over the original $G$ rollouts, with the indicator setting advantages of the deleted rollouts to zero. To solve this argmax, we start with all rollouts and greedily remove those with the most negative estimated contributions, until the constraint in \eqref{eq:cov-contribution-selection} is satisfied. In practice, we length-normalize log-probabilities for both filtering and recentering to avoid direct length bias.

\subsection{Deleting Only Positive-Advantage Rollouts}
\label{sec:positive-only}

Section~\ref{sec:cov-deletion} shows that both high-probability positive-advantage and low-probability negative-advantage rollouts can decrease entropy. That result, however, uses a KL-proximal update whose logit coordinates contain old-policy-dependent scaling. Under ordinary gradient updates with a shared learning rate, their entropy contributions generally differ even when their $h_g$ values are equal. To compare these two classes under the latter update, we additionally analyze one gradient-ascent step on independent sequence logits at the starting point.

\paragraph{Theoretical motivation: conditional entropy effects of positive and negative rollouts.}
For comparison, we consider gradient updates of logits $z_i$ and explicitly require the same learning rate for every logit coordinate. With $L(z)=G^{-1}\sum_gA^{(g)}\log[\operatorname{softmax}(z)]_{i(g)}$, where $i(g)$ is the sequence-space index of sampled rollout $g$, the update is:
\begin{equation}
z_i(\alpha)=z_i+\alpha\frac{\partial L(z)}{\partial z_i},\qquad \alpha>0.
\label{eq:logit-update}
\end{equation}
Here $\alpha$ is the learning rate shared by all logit coordinates: it is the multiplier of each coordinate's gradient that is shared, not the logit or probability change of every sequence. Under this relatively more realistic assumption of a shared learning rate, we obtain the following result.

\begin{proposition}[Greater Entropy Impact of Positive-Advantage Rollouts]
\label{prop:positive-dominance}
In the independent-logit model above, assume $\sum_gA^{(g)}=0$. Here $h_g$ denotes rollout $g$'s additive first-order entropy contribution under the KL-proximal update in Proposition~\ref{prop:single-rollout}, and $h_g^{\mathrm{SGD}}$ denotes its corresponding contribution under the logit gradient update above.
Consider sampled rollouts $\mathbf y_+$ and $\mathbf y_-$ with advantages $A_\pm$ and surprisals $s_\pm=-\log\pi_{\old}(\mathbf y_\pm\mid x)$. Suppose $A_+>0>A_-$ and $s_+<H(\pi_{\old})<s_-$. If their entropy contributions in Proposition~\ref{prop:single-rollout} are equal, $h_+=h_-<0$, then the high-probability rollout contributes more to entropy decrease to first order in a single logit gradient update:
\begin{equation}
\frac{|h_+^{\mathrm{SGD}}|}{|h_-^{\mathrm{SGD}}|}=\frac{\pi_{\old}(\mathbf y_+\mid x)}{\pi_{\old}(\mathbf y_-\mid x)}=\exp(s_--s_+)>1.
\label{eq:positive-dominance}
\end{equation}
\end{proposition}

Thus, even when the two rollouts have equal entropy contributions in Proposition~\ref{prop:single-rollout}, the high-probability positive-advantage rollout makes a larger first-order contribution to entropy reduction in a single logit update, strengthening the tendency toward entropy collapse (see Appendix~\ref{app:positive-only} for the proof). This conditional comparison provides additional motivation for preferentially filtering high-probability positive-advantage candidates.

\paragraph{Method: final deletion rule.}
Based on the above analysis, considering that deleting data itself can negatively reduce data informativeness and potentially degrade accuracy to some extent, we aim to minimize perturbations to the original training process and preserve the original algorithm and data distribution as much as possible. Guided by the preceding theory, we therefore delete only positive-advantage rollouts while retaining negative-advantage ones. Starting from the retained set $\mathcal I_{\mathrm{keep}}$ in Section~\ref{sec:cov-deletion}, we restore any excluded rollouts with nonpositive advantages:
\begin{equation}
\mathcal I_{\mathrm{keep}}\leftarrow\mathcal I_{\mathrm{keep}}\cup\{g\in\{1,\ldots,G\}:A^{(g)}\le0\}.
\label{eq:positive-filter}
\end{equation}

\subsection{Recentering After Deletion and the Full Procedure}
\label{sec:renorm}

Selective deletion changes the composition of retained samples, may make their distribution depart from the old policy, and disrupts the original centering of group advantages. This motivates recentering the retained advantages after deletion.

\paragraph{Theoretical motivation: normalization drift after deletion.}
Generalization can legitimately increase the probability of unsampled solutions similar to rewarded rollouts. Beyond generalization, a normalization offset that does not depend on similarity can also change the probabilities of unsampled paths, which is what we aim to avoid. However, unrecentered advantages and the resulting perturbed distribution tend to exacerbate this issue. To isolate it, we assume i.i.d. sampling of the original rollout group for a fixed prompt and, conditional on the realized group and retained set, analyze an auxiliary distribution-space update without transfer through shared parameters. Using the retained set $\mathcal I_{\mathrm{keep}}$, define the accumulated post-deletion advantage as $\tilde A_i^{\mathrm{keep}}=\sum_{g\in\mathcal I_{\mathrm{keep}}}A^{(g)}\mathbf 1\{\mathbf y_i=\mathbf y^{(g)}\}$. We then obtain the following result:

\begin{proposition}[Invariance condition for zero-advantage sequences]
\label{prop:outside-paths}
Fix a prompt $x$ and a realized group of rollouts sampled i.i.d. from $\pi_{\old}(\cdot\mid x)$. Let $U=\{i:\pi_{\old}(\mathbf y_i\mid x)>0,\ \tilde A_i^{\mathrm{keep}}=0\}$ be the set of unsampled and other zero-advantage paths in the old-policy support. Evaluate $\pi_\theta$ along the auxiliary retained-sample proximal update specified in Appendix~\ref{app:outside-paths}, with $\theta$ varying with $\eta$ and $\pi_\theta=\pi_{\old}$ at $\eta=0$. For every $i\in U$,
\begin{equation}
\frac{d}{d\eta}\log\pi_\theta(\mathbf y_i\mid x)\Bigr|_{\eta=0}
=-\mathbb E_{\pi_{\old}}[\tilde A^{\mathrm{keep}}].
\label{eq:outside-prob-effect}
\end{equation}
\end{proposition}

This proposition shows that all zero-advantage paths share the same first-order log-probability shift: regardless of their quality, their probabilities decrease when the old-policy-weighted mean advantage is positive and increase when it is negative. This shift expresses no preference for more promising paths, and its first-order strength is determined by $\vert{}\mathbb E_{\pi_{\old}}[\tilde A^{\mathrm{keep}}]\vert{}$. Deletion changes the retained score and its old-policy-weighted mean, making such unwarranted probability increases and decreases more pronounced, thereby leading to training instability. The proof is in Appendix~\ref{app:outside-paths}.

\paragraph{Method: probability-weighted recentering.}
Based on the above analysis, we recenter a nonempty retained set as follows. Recomputing $\tilde A^{\mathrm{keep}}$ with the recentered advantages yields a zero old-policy-weighted mean and cancels the first-order normalization shift in the proximal model of Proposition~\ref{prop:outside-paths}:
\begin{equation}
b_{\mathrm{keep}}=\frac{\sum_{g\in\mathcal I_{\mathrm{keep}}}e^{-s_g}A^{(g)}}
{\sum_{g\in\mathcal I_{\mathrm{keep}}}e^{-s_g}},\qquad
\bar A_{\mathrm{keep}}^{(g)}=A^{(g)}-b_{\mathrm{keep}},\quad g\in\mathcal I_{\mathrm{keep}}.
\label{eq:selected-renorm}
\end{equation}
This local analysis motivates the recentering rule; its effectiveness in training is evaluated empirically in Section~\ref{sec:ablation}. Note that this probability weighting is introduced precisely because our data deletion is biased, which violates the i.i.d. assumption.

\paragraph{GRPODropout procedure.}
Based on the above theory and analysis, we present the full \ours{} procedure. To minimize changes to the original training process, we use the selected rollouts and recentered advantages in the original loss, preserving its formulation. The procedure reuses existing rewards and log-probabilities and requires no value network, adding negligible computational overhead for filtering and recentering.

\begin{tcolorbox}[breakable,title={GRPODropout}]
\textbf{Input:} prompt $x$, old policy $\pi_{\old}$, sample count $G$, and the original RL loss.\\
\textbf{1.} Sample $G$ rollouts and compute rewards, native advantages $A^{(g)}$, and old-policy rollout surprisals $s_g$; set advantages to zero if the reward standard deviation is zero.\\
\textbf{2.} Estimate $H(\pi_{\old})$ by $\bar s$, compute and fix $h_g$ using \eqref{eq:single-contribution}, then solve \eqref{eq:cov-contribution-selection} by removing the most negative estimated contributions first to obtain $\mathcal I_{\mathrm{keep}}$.\\
\textbf{3.} Apply \eqref{eq:positive-filter} to restore excluded rollouts with nonpositive native advantages to $\mathcal I_{\mathrm{keep}}$.\\
\textbf{4.} For the retained set, compute and subtract $b_{\mathrm{keep}}$ using \eqref{eq:selected-renorm} to recentered advantages.\\
\textbf{5.} Update the policy using the original RL loss on retained rollouts.
\end{tcolorbox}

\section{Experiments}

We compare \ours{}, original GRPO, and other entropy-aware baselines across three model sizes (1.7B, 4B, and 8B), trained on mathematics with matched prompts. Evaluation spans ten benchmarks covering mathematical reasoning and general QA, with the latter assessing generalization beyond mathematics. Given their small test sets, AMC23/AIME24/AIME25/HMMT25 use Mean@32; the others use Mean@1. Full settings, baselines, and metric conventions appear in Appendix~\ref{app:experimental-setup}.

\subsection{Main Results}
\label{sec:main-results}

\begin{table}[t]
\centering
\caption{Main results on three backbones. Avg.\ averages ten benchmarks; Base denotes the untrained backbone. Bold marks the best result within each backbone block.}
\label{tab:main}
\small
\setlength{\tabcolsep}{1.5pt}
\scalebox{0.892}{%
\begin{tabular}{clccccccccccc}
\toprule
& \multirow{2}{*}{Method} & \multicolumn{7}{c}{Mathematical reasoning} & \multicolumn{3}{c}{General QA} & \multirow{2}{*}{Avg.} \\
\cmidrule(lr){3-9}\cmidrule(lr){10-12}
& & GSM8k & Math500 & AMC23 & Olympiad & AIME24 & AIME25 & HMMT25 & ARC-C & MMLUP & SGPQA & \\
\midrule
\multirow{7}{*}{\rotatebox[origin=c]{90}{\textbf{Qwen3-1.7B}}}
 & Base        & 89.92 & 80.60 & 59.92 & 42.96 & 22.08 & 21.04 & 11.46 & 87.37 & 57.01 & 29.28 & 50.16 \\
 & GRPO        & 90.60 & 84.00 & 72.27 & 51.56 & 28.54 & 23.44 & 14.06 & 87.80 & \textbf{57.91} & \textbf{31.07} & 54.13 \\
 & DAPO        & 88.86 & 85.20 & 72.27 & 52.00 & \textbf{33.75} & \textbf{26.15} & 15.21 & 87.63 & 55.62 & 29.70 & 54.64 \\
 & Clip-Cov    & 82.34 & 77.20 & 53.91 & 36.15 & 11.56 & 12.71 & 5.21 & 87.37 & 55.19 & 28.26 & 44.99 \\
 & KL-Cov      & 90.07 & \textbf{86.40} & 70.39 & 50.07 & 31.67 & 25.73 & 14.79 & \textbf{88.31} & 56.60 & 29.68 & 54.37 \\
 & Beyond8020  & 89.01 & 84.00 & 72.19 & 50.22 & 29.48 & 25.00 & \textbf{18.02} & 86.69 & 56.03 & 29.78 & 54.04 \\
 & GRPODropout & \textbf{90.83} & 85.60 & \textbf{75.55} & \textbf{52.15} & 31.46 & 25.62 & 15.83 & 86.35 & 56.64 & 30.02 & \textbf{55.01} \\
\midrule
\multirow{7}{*}{\rotatebox[origin=c]{90}{\textbf{Qwen3-4B}}}
 & Base        & 94.77 & 85.40 & 72.50 & 47.41 & 37.92 & 26.25 & 14.27 & 94.20 & 69.19 & 39.85 & 58.18 \\
 & GRPO        & 94.77 & 91.80 & 89.45 & 58.37 & 42.60 & 33.23 & 18.96 & 94.11 & 68.35 & 38.64 & 63.03 \\
 & DAPO        & 94.47 & 89.80 & 86.72 & 57.48 & 42.40 & 36.88 & 19.06 & 94.11 & 67.48 & 38.52 & 62.69 \\
 & Clip-Cov    & 94.54 & \textbf{92.00} & 86.33 & 59.41 & 54.06 & 41.56 & 27.08 & 94.45 & 69.32 & 40.79 & 65.95 \\
 & KL-Cov      & \textbf{94.92} & \textbf{92.00} & 87.89 & 59.11 & \textbf{54.79} & 40.83 & \textbf{27.60} & 94.03 & 69.27 & \textbf{40.83} & \textbf{66.13} \\
 & Beyond8020  & 94.39 & 91.60 & \textbf{89.92} & \textbf{60.00} & 51.88 & 41.25 & 23.65 & \textbf{94.54} & 68.21 & 39.39 & 65.48 \\
 & GRPODropout & 94.77 & 91.80 & 88.28 & 58.96 & 51.98 & \textbf{43.02} & 25.94 & 94.11 & \textbf{69.35} & 40.25 & 65.85 \\
\midrule
\multirow{7}{*}{\rotatebox[origin=c]{90}{\textbf{Qwen3-8B}}}
 & Base        & 95.45 & 82.80 & 67.50 & 46.07 & 37.40 & 24.69 & 11.98 & 95.90 & 73.14 & 43.61 & 57.85 \\
 & GRPO        & 95.30 & 91.20 & 85.86 & 59.11 & 53.54 & 38.44 & 24.38 & 95.73 & \textbf{74.22} & \textbf{45.54} & 66.33 \\
 & DAPO        & 95.53 & 92.20 & \textbf{90.31} & \textbf{61.33} & 52.40 & 38.85 & 21.25 & 95.65 & 72.39 & 43.75 & 66.37 \\
 & Clip-Cov    & 95.38 & 91.20 & 84.77 & 60.44 & 54.90 & 42.08 & \textbf{27.60} & 95.56 & 73.76 & 45.05 & 67.07 \\
 & KL-Cov & \textbf{95.75} & \textbf{93.20} & 86.72 & \textbf{61.33} & 54.17 & 40.31 & 27.50  & 95.48 & 73.81 & 44.88 & \textbf{67.32} \\
 & Beyond8020  & 95.45 & 92.20 & 87.58 & 60.59 & \textbf{57.29} & 39.06 & 24.69 & \textbf{95.99} & 73.23 & 44.57 & 67.07 \\
 & GRPODropout & 95.53 & 92.20 & 87.66 & 60.74 & 55.31 & \textbf{42.81} & 25.21 & 94.97 & 73.50 & 44.99 & 67.29 \\
\bottomrule
\end{tabular}%
}
\end{table}

\begin{figure}[t]
\centering
\includegraphics[width=\linewidth]{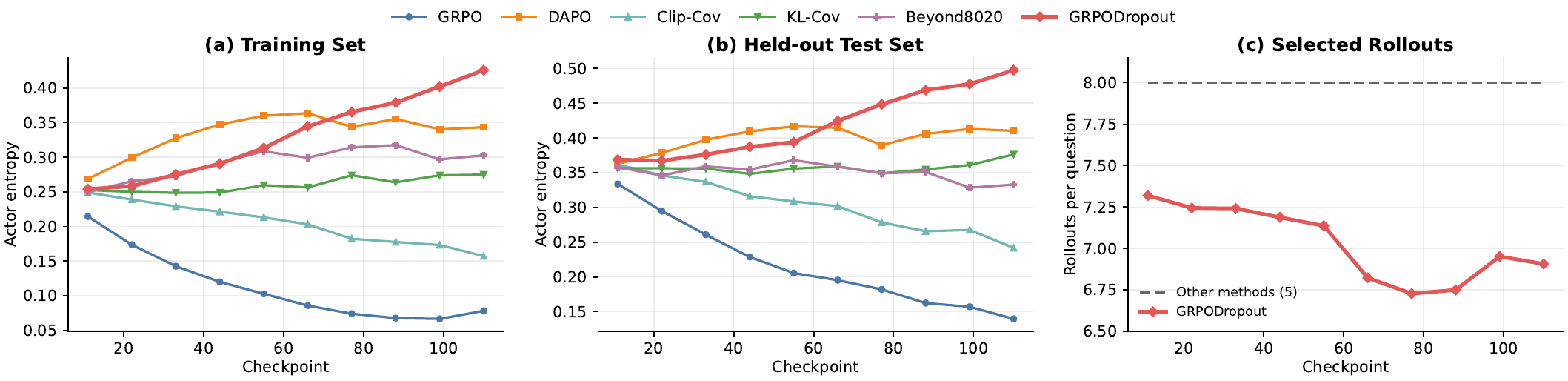}
\caption{Qwen3-4B actor entropy and rollout usage across checkpoints. (a--b) Token-mean entropy on training DAPO-17K and held-out MMLU-Pro (MMLUP); (c) mean update rollouts per question in training batches, excluding all-zero-advantage groups. The other five methods use 8 rollouts at every recorded checkpoint (gray dashed line). Curves connect raw observations without smoothing.}
\label{fig:entropy-rollouts}
\end{figure}

\paragraph{GRPODropout targets rollouts that may reduce policy entropy.}
Under the same rollout sampling budget, we assess whether using fewer retained rollouts in policy updates can preserve or improve performance over original GRPO. Table~\ref{tab:main} reports results on three model sizes: selective deletion with recentering improves average accuracy over GRPO by 1.55 percentage points on average. These results support our data-usage insight that using all sampled rollouts is not always the most effective training strategy, and that selectively removing a subset can instead yield better performance. GRPODropout also exceeds KL-Cov, the strongest entropy-aware baseline by average score, by 0.11 points on average. Finally, our method is orthogonal in design to these optimization methods, offering both a data-selection insight and a complementary strategy for mitigating entropy collapse.

\paragraph{Removing few rollouts substantially mitigates entropy collapse.}
We seek to preserve performance and exploration with minimal changes to training. Because sequence-level entropy is strongly affected by response length, we follow prior work~\citep{cui2025entropymechanism} and report token-level conditional entropy averaged over valid response tokens. We measure this entropy as $-\sum_v\pi_\theta(v\mid x,\mathbf y_{<t})\log\pi_\theta(v\mid x,\mathbf y_{<t})$, where $v$ ranges over the vocabulary. Figure~\ref{fig:entropy-rollouts} tracks this metric for Qwen3-4B on DAPO-17K and held-out MMLU-Pro. While GRPO exhibits entropy collapse, GRPODropout achieves 446.1\% and 256.0\% higher entropy, respectively, at the end of training. Across the ten recorded checkpoints, excluding all-zero-advantage groups, it retains 7.03 rollouts per question for the update—12.1\% fewer than GRPO's baseline of eight. Notably, compared to Beyond8020, which masks out 80\% of tokens, our method removes a mere 12.1\% of the data while achieving comparable or even superior policy entropy and performance, demonstrating the efficacy and value of rollout-level deletion.

\subsection{Ablation Studies}
\label{sec:ablation}

Our method has two main components: rollout deletion and advantage recentering. We ablate the effects of deleting positive-advantage (Pos.) and negative-advantage (Neg.) rollouts, enabling recentering (Center), and using probability weighting during recentering (P-wt.). Specifically, with Center/P-wt. set to $\checkmark/\checkmark$, we subtract the probability-weighted mean advantage as in \eqref{eq:selected-renorm}; with $\checkmark/\times$, we subtract the equal-weight mean of retained advantages; with $\times/\times$, no recentering is performed. The first row of each model block reports original GRPO without rollout deletion. Table~\ref{tab:ablation-4b} reports results on Qwen3-1.7B and Qwen3-4B.

\begin{table}[t]
\centering
\caption{Qwen3-1.7B and Qwen3-4B ablations of positive- and negative-advantage rollouts, advantage recentering, and old-policy probability weighting of the mean. $\checkmark$/$\times$: enabled/disabled; --: not applicable. Avg.\ Ent reports benchmark-average actor entropy.}
\label{tab:ablation-4b}
\setlength{\tabcolsep}{1.5pt}
\scalebox{0.733}{%
\begin{tabular}{@{}cccccccccccccccc@{}}
\toprule
\multicolumn{2}{c}{Deletion} & \multicolumn{2}{c}{Advantages} & \multicolumn{7}{c}{Mathematical reasoning} & \multicolumn{3}{c}{General QA} & \multirow{2}{*}{Avg.} & \multirow{2}{*}{Avg.\ Ent} \\
\cmidrule(lr){1-2}\cmidrule(lr){3-4}\cmidrule(lr){5-11}\cmidrule(lr){12-14}
Pos. & Neg. & Center & P-wt. & GSM8k & Math500 & AMC23 & Olympiad & AIME24 & AIME25 & HMMT25 & ARC-C & MMLUP & SGPQA & & \\
\midrule
\multicolumn{16}{c}{\textbf{Qwen3-1.7B}} \\
\midrule
$\times$ & $\times$ & -- & -- & 90.60 & 84.00 & 72.27 & 51.56 & 28.54 & 23.44 & 14.06 & 87.80 & \textbf{57.91} & \textbf{31.07} & 54.13 & 0.0503 \\
$\checkmark$ & $\checkmark$ & $\checkmark$ & $\checkmark$ & 90.30 & \textbf{86.80} & 70.47 & 49.04 & 30.94 & 24.90 & 13.75 & 87.71 & 56.53 & 29.77 & 54.02 & 0.2025 \\
$\checkmark$ & $\times$ & $\times$ & $\times$ & 77.33 & 39.00 & 10.70 & 8.74 & 0.00 & 0.10 & 0.00 & 87.12 & 44.04 & 21.09 & 28.81 & 0.3282 \\
$\checkmark$ & $\times$ & $\checkmark$ & $\times$ & 89.76 & 85.00 & 65.31 & 46.67 & 27.29 & 22.40 & 11.46 & \textbf{87.88} & 56.64 & 29.44 & 52.19 & 0.2239 \\
$\checkmark$ & $\times$ & $\checkmark$ & $\checkmark$ & \textbf{90.83} & 85.60 & \textbf{75.55} & \textbf{52.15} & \textbf{31.46} & \textbf{25.62} & \textbf{15.83} & 86.35 & 56.64 & 30.02 & \textbf{55.01} & 0.1739 \\
\midrule
\multicolumn{16}{c}{\textbf{Qwen3-4B}} \\
\midrule
$\times$ & $\times$ & -- & -- & \textbf{94.77} & 91.80 & \textbf{89.45} & 58.37 & 42.60 & 33.23 & 18.96 & 94.11 & 68.35 & 38.64 & 63.03 & 0.0528 \\
$\checkmark$ & $\checkmark$ & $\checkmark$ & $\checkmark$ & \textbf{94.77} & 91.00 & 87.03 & 58.22 & 51.46 & 41.25 & 25.00 & 93.34 & 69.09 & 40.10 & 65.13 & 0.2425 \\
$\times$ & $\checkmark$ & $\checkmark$ & $\checkmark$ & 94.54 & 91.40 & 87.97 & \textbf{59.26} & 51.04 & 41.88 & 23.44 & \textbf{94.28} & 68.80 & 40.43 & 65.30 & 0.2043 \\
$\checkmark$ & $\times$ & $\times$ & $\times$ & 0.38 & 0.20 & 2.19 & 0.00 & 0.00 & 0.00 & 0.00 & 88.57 & 42.35 & 26.34 & 16.00 & 0.0200 \\
$\checkmark$ & $\times$ & $\checkmark$ & $\times$ & 94.62 & \textbf{92.20} & 89.30 & 58.22 & \textbf{52.81} & 42.08 & 25.10 & 93.94 & \textbf{69.48} & \textbf{40.44} & 65.82 & 0.2174 \\
$\checkmark$ & $\times$ & $\checkmark$ & $\checkmark$ & \textbf{94.77} & 91.80 & 88.28 & 58.96 & 51.98 & \textbf{43.02} & \textbf{25.94} & 94.11 & 69.35 & 40.25 & \textbf{65.85} & 0.2140 \\
\bottomrule
\end{tabular}%
}
\end{table}

\paragraph{Deleting only positive-advantage rollouts performs best.}
With probability-weighted recentering fixed, we compare deleting positive- and/or negative-advantage candidates. Deletion generally increases entropy and improves final accuracy in most settings, while the magnitude of the accuracy gain depends on the deletion pattern and recentering scheme. Deleting both positive and negative candidates yields the largest entropy gain but incurs a slight accuracy loss, showing that data removal can sacrifice performance and supporting our motivation to delete as little data as possible. Deleting only positive candidates achieves higher entropy and accuracy than deleting only negative candidates, consistent with our theoretical result that positive-advantage rollouts have a greater effect on entropy reduction.

\paragraph{Recentering is necessary.}
With deletion restricted to positive-advantage candidates, removing recentering causes severe performance degradation and training collapse, demonstrating the importance of recentering in the tested deletion configurations. We further introduce probability weighting to calibrate the retained advantages under the old-policy measure. On Qwen3-4B, probability weighting has little effect, improving the average score by only 0.03 points over equal-weight recentering. Its benefit is more pronounced on Qwen3-1.7B, where removing the weighting lowers the average score by 2.82 points, suggesting that probability weighting can sometimes improve training stability.

\subsection{More Precise Exploration and More Diverse Reasoning Paths}
\label{sec:precision-exploration}

The full GRPODropout method increases entropy while largely preserving or improving average accuracy across the three model sizes. We further examine the value of this retained entropy from two perspectives: exploration precision, to assess accuracy preservation; and reasoning-path diversity, to evaluate the preservation of exploration capacity.

\paragraph{Metrics.}
We first assess how accurately model sampling produces correct solutions while limiting repeated incorrect final answers. We compare all methods on Qwen3-4B across AMC23, AIME24, AIME25, and HMMT25, sampling 32 rollouts per question. Pass@32 is computed per question within each benchmark and then averaged equally across AMC23, AIME24, AIME25, and HMMT25. For Table~\ref{tab:precision-exploration}, Entropy is token-mean actor entropy over valid response tokens at the final checkpoint, averaged equally across the four benchmarks. Inspired by the answer-consistency perspective of~\citet{wang2022selfconsistency}, we define answer-string dispersion $W$ as the number of distinct nonempty final-answer strings among a question's incorrect responses. $W$ measures answer-string dispersion among incorrect final answers and can reveal recurring answer patterns. To distinguish relatively easier and harder questions, we use a fixed cross-method partition: a question enters the $\geq 8$ subset only if every one of the six systems has at least eight correct responses, and enters the $<8$ subset only if every system has fewer than eight. We average $W$ within each subset.

We next assess whether models approach the same problem through diverse reasoning paths. To this end, we use GLM-5.2 to summarize each response's reasoning path in three to five sentences, excluding abandoned reasoning, and then encode the summaries with bge-m3~\citep{chen-etal-2024-m3}. Following~\citet{kirk2023diversity,chen-etal-2026-dra}, Summary distance measures the mean pairwise cosine distance between summary embeddings for the same question, reflecting semantic separation. Following~\citet{friedman2023vendi,cui2026single}, Vendi measures effective diversity as the exponential of the entropy of the eigenvalues of the Gram matrix normalized by the number of summaries, accounting for similarity and repetition. Higher values indicate greater diversity. Here, $W$ measures diversity among incorrect final answers, whereas Summary distance and Vendi measure diversity in the summarized reasoning paths, including paths that may lead to the same final answer. Full prompt and run settings are given in Appendix~\ref{app:table3-diversity}. 

\paragraph{A favorable trade-off between correctness and diversity.}
The results show that our method achieves the best or second-best reported mean on all eight metrics and ranks first among the six systems, with an overall mean rank of 1.63 under equal weighting. In particular, our method attains the highest reasoning-path diversity on the $\geq 8$ subset and the highest actor entropy while achieving lower $W$ than all baselines except KL-Cov. Compared with GRPO, higher solution coverage and actor entropy accompany fewer distinct incorrect answers and higher values on all four path-diversity measures, demonstrating that our improvement in Pass@1 accuracy does not come at the expense of Pass@32 or the model's exploratory capability, but rather enhances them simultaneously. Notably, although DAPO exhibits relatively high entropy and reasoning-path diversity on the $\geq 8$ subset, this does not translate into the highest Pass@32, indicating that it sacrifices accuracy for exploratory diversity. Correspondingly, our method maximally preserves exploration accuracy—namely, higher Pass@1 and Pass@32, and fewer distinct incorrect answers—while fully maintaining the path diversity and entropy of correct answers, ensuring no loss of the model's exploratory capability and achieving a simultaneous preservation of accuracy and entropy.

\begin{table}[t]
\centering
\small
\caption{Qwen3-4B solution coverage, wrong-answer dispersion, and reasoning diversity. Definitions are in Section~\ref{sec:precision-exploration}. Bold marks the best and underline marks the second-best in each column.}
\label{tab:precision-exploration}
\setlength{\tabcolsep}{3pt}
\scalebox{0.9}{%
\begin{tabular}{lccccccccc}
\toprule
\multirow{2}{*}{Method}
& \multicolumn{3}{c}{Correctness}
& \multicolumn{4}{c}{Path diversity}
& \multirow{2}{*}{Entropy $\uparrow$}
& \multirow{2}{*}{\shortstack{Overall\\mean rank $\downarrow$}} \\
\cmidrule(lr){2-4}\cmidrule(lr){5-8}
& pass@32 $\uparrow$
& $W_{\geq 8}$ $\downarrow$
& $W_{<8}$ $\downarrow$
& Dist.$_{\geq 8}$ $\uparrow$
& Dist.$_{<8}$ $\uparrow$
& Vendi$_{\geq 8}$ $\uparrow$
& Vendi$_{<8}$ $\uparrow$
& & \\
\midrule
GRPO
& 63.54 & 1.1846 & 7.6216
& 0.1063 & 0.1641 & 1.8311 & 2.4194
& 0.0448 & 5.63 \\
DAPO
& 69.38 & 1.6154 & 8.0811
& \underline{0.1150} & 0.1798
& \underline{1.9080} & 2.6013
& \underline{0.1796} & 3.88 \\
Clip-Cov
& \textbf{76.88} & 0.7692 & 5.2162
& 0.1139 & \textbf{0.2141}
& 1.9041 & \textbf{2.9851}
& 0.0900 & \underline{2.63} \\
KL-Cov
& 72.71 & \textbf{0.4615} & \textbf{3.0000}
& 0.1134 & 0.2014 & 1.8930 & 2.8510
& 0.1509 & 3.25 \\
Beyond8020
& 73.54 & 1.1231 & 7.8378
& 0.1142 & 0.1726 & 1.9037 & 2.5252
& 0.1671 & 4.00 \\
GRPODropout
& \underline{75.21}
& \underline{0.4923} & \underline{4.2432}
& \textbf{0.1151} & \underline{0.2090}
& \textbf{1.9114} & \underline{2.8992}
& \textbf{0.2291} & \textbf{1.63} \\
\bottomrule
\end{tabular}}
\end{table}

\section{Conclusion}

We find that some high-probability positive rollouts can hinder training in reinforcement learning algorithms such as GRPO. Removing these rollouts improves performance and mitigates entropy collapse. Based on this finding, we propose GRPODropout, which selectively removes high-probability positive rollouts and recenters the retained advantages to stabilize training, while leaving the original RL loss unchanged. We further extend the theoretical analysis of entropy dynamics from tokens to complete rollouts, providing theoretical motivation for our rollout-deletion strategy. Extensive experiments across three model scales and ten benchmarks demonstrate that GRPODropout outperforms the original GRPO while using fewer retained rollouts in policy updates.

\paragraph{Limitations.} First, GRPODropout does not outperform the strongest baseline at every model scale. It is worth pointing out that we do not merely provide an entropy collapse mitigation method; rather, our work includes the insight that reinforcing some rollouts—even correct ones—can be counterproductive, which also possesses clear value. Second, given the complexity of RL training, our theory itself cannot fully explain our method. In fact, the primary role of the theory is to serve as motivation for method design, acting as the initial starting point. The final method is validated through its advantages over the original GRPO and the ablation study results. Finally, GRPODropout relies on within-prompt differences in advantages and probabilities across multiple rollouts. It therefore does not directly apply to single-rollout training, while its effectiveness with very small groups remains to be evaluated.

\bibliography{custom}
\bibliographystyle{iclr2027_conference}

\appendix
\newpage
\section{Supplementary Material}
\label{app:supplementary}

\subsection{Proofs of Proposition~\ref{prop:single-rollout} and Proposition~\ref{prop:entropy-covariance}}
\label{app:proofs}

\paragraph{Local update approximation.}
To extend token-level entropy analysis to complete rollouts, we adopt a distribution-space KL-proximal policy-improvement model following~\citet{abdolmaleki2018maximum}. For a candidate policy $\pi(\cdot\mid x)$, we define its proximal update as
\begin{equation}
L_A(\pi)=\frac1G\sum_{g=1}^G A^{(g)}
\frac{\pi(\mathbf y^{(g)}\mid x)}{\pi_{\old}(\mathbf y^{(g)}\mid x)},\quad
\pi^{(\eta)}=\arg\max_{\pi\in\Delta_{\old}}
\Bigl\{L_A(\pi)-\frac1\eta\mathrm{KL}\!\bigl(\pi\,\|\,\pi_{\old}\bigr)\Bigr\}.
\tag{A.1}\label{eq:local-update}
\end{equation}
Here $\eta>0$ controls the size of the proximal update, and $\Delta_{\old}$ contains normalized distributions over complete sequences supported on $\{\mathbf y_i:\pi_{\old}(\mathbf y_i\mid x)>0\}$. \eqref{eq:local-update} specifies an idealized update directly over sequence distributions. Its old-policy KL penalty specifies the update geometry, while the KL regularizer relative to the reference policy is omitted. We analyze the update around $\eta=0$, taking $\pi^{(0)}=\pi_{\old}$, with the sampled group and its finite advantages held fixed. We assume $H(\pi_{\old})<\infty$ and use rollout advantages without length normalization. We derive the sequence-score representation and closed-form update below.

\paragraph{Notation for the proofs.}
For the fixed prompt $x$, we use the shorthand $p(\mathbf y)=\pi_{\old}(\mathbf y\mid x)$ and $q(\mathbf y)=\pi(\mathbf y\mid x)$ for the old and candidate policies, with $p_i=p(\mathbf y_i)$ and $q_i=q(\mathbf y_i)$. The proximal policies are written $q_i^{(\eta)}=\pi^{(\eta)}(\mathbf y_i\mid x)$ and $q_{\mathrm{keep},i}^{(\eta)}=\pi_\theta(\mathbf y_i\mid x)$, where in the latter case $\theta$ varies with $\eta$ along the retained-sample proximal update. These abbreviations are used only in this appendix.

\paragraph{Sampling-aligned sequence score.}
With the sampled group and its advantages fixed, express the sampled surrogate in sequence space using
\begin{equation}
\tilde A_i=\frac{1}{G p_i}\sum_{g=1}^G A^{(g)}\mathbf 1\{\mathbf y_i=\mathbf y^{(g)}\},\qquad p_i>0.
\tag{A.2}\label{eq:aligned-score}
\end{equation}
Here $\mathbf 1\{\cdot\}$ is the indicator function, and sums over sequence indices are restricted to the old-policy support. Repeated samples of the same sequence contribute additively, while a sequence not sampled in the current group has $\tilde A_i=0$ and receives no direct update signal. Indeed,
\[
\sum_iq_i\tilde A_i
=\sum_iq_i\frac{1}{G p_i}\sum_{g=1}^G A^{(g)}\mathbf 1\{\mathbf y_i=\mathbf y^{(g)}\}
=\frac1G\sum_{g=1}^G A^{(g)}\frac{q(\mathbf y^{(g)})}{p(\mathbf y^{(g)})}=L_A(q).
\]
Intuitively, $\tilde A_i$ measures the marginal value of assigning probability to sequence $\mathbf y_i$: transferring a small probability mass $\delta$ from $\mathbf y_j$ to $\mathbf y_i$ changes $L_A$ by $\delta(\tilde A_i-\tilde A_j)$.
Consequently, at the starting point,
$\nabla_\theta L_A=G^{-1}\sum_gA^{(g)}\nabla_\theta\log\pi_\theta(\mathbf y^{(g)}\mid x)|_{\theta=\theta_{\old}}$; the old policy and denominator are held fixed. Omitting $1/p_i$ from \eqref{eq:aligned-score} would change the coefficient of sample $g$ to $e^{-s_g}A^{(g)}/G$, which cannot generally be corrected by one global learning rate. The inverse-probability factor is needed only for the mathematical correspondence and is not explicitly computed by the algorithm; its difference from a direct logit update is analyzed in Appendix~\ref{app:positive-only}.

\paragraph{Setup and regularity conditions.}
Fix a prompt $x$ and consider only the positive-probability support of the old policy, $\{\mathbf y_i:p_i>0\}$. Assume that this set is finite or countable, $H(p)<\infty$, the sample count $G$ is finite, and every advantage $A^{(g)}$ is a finite real number; samples and advantages are fixed throughout the update. Since the score in \eqref{eq:aligned-score} can be nonzero only on finitely many sampled rollouts and each has $p_i>0$, the empirical score is bounded. Therefore $Z(\eta)=\sum_i p_i e^{\eta\tilde A_i}$ exists and is positive for every finite $\eta$. In a neighborhood of zero, $q_i^{(\eta)}/p_i$ and its derivative with respect to $\eta$ are uniformly bounded, and each summand of the entropy derivative is dominated by a constant multiple of $p_i(1+|\log p_i|)$. Interchanging differentiation and summation below is consequently valid. This argument is conditional on a fixed finite sampled group and does not mistake a finite first moment for exponential integrability of an arbitrary unbounded score. Taking a further expectation over prompts requires the corresponding additional integrability conditions.

\paragraph{Closed-form KL-proximal update.}
The stationarity condition for \eqref{eq:local-update} under normalization gives $q_i\propto p_i\exp(\eta\tilde A_i)$. Normalizing over the old-policy support yields
\begin{equation}
\pi^{(\eta)}(\mathbf y_i\mid x)=
\frac{\pi_{\old}(\mathbf y_i\mid x)\exp(\eta\tilde A_i)}
{\sum_j\pi_{\old}(\mathbf y_j\mid x)\exp(\eta\tilde A_j)}.
\tag{A.3a}\label{eq:closed-form-update}
\end{equation}
The updated policy converges to $\pi_{\old}$ as $\eta\to0^+$, so we extend the update continuously to $\pi^{(0)}=\pi_{\old}$. Equivalently, using the shorthand defined above,
\begin{equation}
q_i^{(\eta)}=\frac{p_i\exp(\eta\tilde A_i)}{Z(\eta)},\qquad Z(\eta)=\sum_jp_j\exp(\eta\tilde A_j).
\tag{A.3b}\label{eq:appendix-q}
\end{equation}
To verify that this is the unique global optimizer rather than merely a stationary point, observe that for every $q\in\Delta_{\old}$,
\[
\sum_iq_i\tilde A_i-\frac1\eta\mathrm{KL}(q\|p)
=\frac1\eta\log Z(\eta)-\frac1\eta\mathrm{KL}(q\|q^{(\eta)}).
\]
By nonnegativity of KL, the right-hand side is maximized exactly at $q=q^{(\eta)}$; a distribution with $\mathrm{KL}(q\|p)=\infty$ cannot improve the objective. This argument applies to finite and countable supports.

Taking logs of \eqref{eq:appendix-q} and differentiating yields
\[
\frac1{q_i^{(\eta)}}\frac{dq_i^{(\eta)}}{d\eta}
=\tilde A_i-\frac{Z'(\eta)}{Z(\eta)}
=\tilde A_i-\mathbb E_{q^{(\eta)}}[\tilde A].
\]
Hence
\begin{equation}
\frac{dq_i^{(\eta)}}{d\eta}=q_i^{(\eta)}\left(\tilde A_i-\mathbb E_{q^{(\eta)}}[\tilde A]\right).
\tag{A.4}\label{eq:q-derivative}
\end{equation}
At $\eta=0$, $q_i^{(0)}=p_i$, and therefore
\begin{equation}
\left.\frac{dq_i^{(\eta)}}{d\eta}\right|_{\eta=0}=p_i\left(\tilde A_i-\mathbb E_p[\tilde A]\right).
\tag{A.5}\label{eq:q-derivative-zero}
\end{equation}
Differentiating $H(q^{(\eta)})=-\sum_iq_i^{(\eta)}\log q_i^{(\eta)}$ gives
\[
\frac{dH(q^{(\eta)})}{d\eta}=-\sum_i\frac{dq_i^{(\eta)}}{d\eta}\left(1+\log q_i^{(\eta)}\right).
\]
Because $\sum_iq_i^{(\eta)}=1$, we have $\sum_i dq_i^{(\eta)}/d\eta=0$, so the constant term vanishes. Therefore,
\[
\frac{dH(q^{(\eta)})}{d\eta}=-\sum_i\frac{dq_i^{(\eta)}}{d\eta}\log q_i^{(\eta)}.
\]
Substituting $\eta=0$ and \eqref{eq:q-derivative-zero} gives
\begin{equation}
\left.\frac{dH(q^{(\eta)})}{d\eta}\right|_{\eta=0}
=-\sum_ip_i\left(\tilde A_i-\mathbb E_p[\tilde A]\right)\log p_i
=-\Cov_p(\tilde A,\log p).
\tag{A.6}\label{eq:appendix-entropy-cov}
\end{equation}
This is the entropy derivative of the reward-driven KL-proximal update specified by \eqref{eq:local-update}; it does not apply to an arbitrary online-RL parameter update.

\paragraph{Proposition~\ref{prop:single-rollout}: per-sample contribution.}
For each sampled index $g$, define the score component $a_i^{[g]}=A^{(g)}\mathbf 1\{\mathbf y_i=\mathbf y^{(g)}\}/(Gp_i)$, so that $\tilde A_i=\sum_ga_i^{[g]}$. Since
\[
\mathbb E_p[a^{[g]}]=\frac{A^{(g)}}G,\qquad
\mathbb E_p[a^{[g]}\log p]=-\frac{A^{(g)}}G s_g,\qquad
\mathbb E_p[\log p]=-H(p),
\]
and \eqref{eq:appendix-entropy-cov} is linear in its score argument,
\begin{equation}
h_g=-\Cov_p(a^{[g]},\log p)
=\frac{A^{(g)}}G\bigl(s_g-H(p)\bigr).
\tag{A.7}\label{eq:appendix-single}
\end{equation}
This proves \eqref{eq:single-contribution}. Throughout, $g$ denotes a sampled index and $i$ an index in the complete-sequence space. This additive decomposition includes the normalization-induced changes of other sequences, not only the entropy term of rollout $g$ itself. Its sign is determined by $A^{(g)}(s_g-H(p))$: a relatively high-probability positive-advantage rollout and a relatively low-probability negative-advantage rollout decrease entropy; the reverse combinations increase it. The contribution is zero if $A^{(g)}=0$ or $s_g=H(p)$.

\paragraph{Proposition~\ref{prop:entropy-covariance}: group entropy derivative.}
Summing \eqref{eq:appendix-single} over sampled indices gives the entropy derivative of $q^{(\eta)}$ at $\eta=0$. Under the zero-sum condition of Proposition~\ref{prop:entropy-covariance}, $\mathbb E_p[\tilde A]=G^{-1}\sum_gA^{(g)}=0$, so
\begin{equation}
\begin{aligned}
\left.\frac{dH(q^{(\eta)})}{d\eta}\right|_{\eta=0}
&=\sum_g h_g=\frac1G\sum_g A^{(g)}s_g\\
&=\frac1G\sum_g A^{(g)}(s_g-\bar s)
=\operatorname{Cov}_{g=1}^{G}\!\bigl(A^{(g)},s_g\bigr).
\end{aligned}
\tag{A.8}\label{eq:group-identity}
\end{equation}
This proves \eqref{eq:covariance-conclusion}, using the same $s_g$ and $\bar s$ as in the main text. Thus a negative empirical covariance means that the KL-proximal update driven by this fixed sampled group decreases entropy at the starting point, while a positive value means that it increases entropy. When the covariance is zero, only the first-order term vanishes; finite-step entropy change is undetermined. \eqref{eq:group-identity} is exact conditional on the current samples and weights, not an entropy guarantee for a neural-network update.

\paragraph{Monte Carlo evaluation and greedy selection.}
For the sequence-level criterion, we estimate $H(p)$ by the sample mean $\bar s$ of the existing $G$ rollouts and substitute it into \eqref{eq:single-contribution}. Each $s_g=-\sum_{t=1}^{T_g}\log\pi_{\old}(y_t^{(g)}\mid x,\mathbf y_{<t}^{(g)})$ sums negative old-policy log-probabilities over valid response tokens; the length-normalized implementation is described below.

The covariance in \eqref{eq:cov-contribution-selection} is computed over the original $G$ rollouts. Since $\sum_{g=1}^G(s_g-\bar s)=0$, centering the masked advantages leaves
\[
\begin{aligned}
\operatorname{Cov}_{g=1}^{G}\!\bigl(A^{(g)}\mathbf 1\{g\in\mathcal I\},s_g\bigr)
&=\frac1G\sum_{g\in\mathcal I} A^{(g)}(s_g-\bar s)\\
&=\sum_{g\in\mathcal I}\left.h_g\right|_{H(p)\to\bar s}.
\end{aligned}
\]
Thus the covariance constraint is exactly equivalent to a nonnegative sum of the remaining MC contributions. Recomputing the covariance on the retained subset would generally change the criterion because its mean surprisal need not equal $\bar s$.

For the greedy-selection argument, use $h_g$ with the MC substitution specified in Section~\ref{sec:cov-deletion}. Start with all $G$ rollout indices. If their total contribution is negative, remove negative contributions in ascending order until the retained sum is nonnegative. Among sets with the same number of removals, removing the most negative contributions produces the largest retained sum, while removing nonnegative contributions cannot help. Thus the first feasible set uses the fewest removals and retains the most rollouts, solving \eqref{eq:cov-contribution-selection}. Removing all negative contributions leaves a nonnegative sum, so a feasible retained set exists. If $\sum_{g=1}^G h_g\ge0$, retaining all $G$ rollouts is optimal. This establishes maximum-cardinality optimality for the fixed MC contributions.

\paragraph{Length normalization in implementation.}
\label{app:length-normalization}
To avoid direct length bias from summing token log-probabilities, the implementation uses the length-normalized surprisal $s_g^{\mathrm{len}}=s_g/T_g$ for both filtering and recentering, with $T_g$ counting valid response tokens. Filtering uses the original-group mean of $s_g^{\mathrm{len}}$, and recentering uses weights proportional to $\exp(-s_g^{\mathrm{len}})=\pi_{\old}(\mathbf y^{(g)}\mid x)^{1/T_g}$, computed by a softmax over retained rollouts. Thus responses with equal mean token log-probability receive equal recentering weights regardless of length. These normalized quantities are implementation proxies: their group mean is a filtering baseline rather than a complete-sequence entropy estimate, and the implemented zero-mean condition is defined by these weights. The theoretical results retain full sequence probabilities as their analysis model.

\paragraph{General objective and probability-ratio surrogate.}
\eqref{eq:local-update} specifies a distribution-space KL-proximal geometry for analyzing the sampled reward update. Its penalty $\eta^{-1}\mathrm{KL}(q\|p)$ limits change relative to the old policy, whereas the reference penalty $\beta\,\mathrm{KL}(q\|\pi_{\mathrm{ref}})$ in the original RL objective penalizes deviation from a fixed reference model. The generalized step size $\eta$ and reference-KL coefficient $\beta$ serve different purposes; the proximal model analyzes the reward contribution separately from clipping, length weights, and reference regularization. For the reward term, subtracting a fixed baseline that depends only on $x$ does not change the expected policy gradient~\citep{williams1992reinforce,sutton1999policy}. For any fixed advantage function $A(x,\mathbf y)$, if the relevant expectation exists and the support of $q$ is contained in that of $p$, the probability-ratio identity gives
\[
\sum_iq_iA(x,\mathbf y_i)=\sum_ip_i\frac{q_i}{p_i}A(x,\mathbf y_i)=\mathbb E_{i\sim p}\!\left[\frac{q_i}{p_i}A(x,\mathbf y_i)\right].
\]
Taking $A^{(g)}=A(x,\mathbf y^{(g)})$ gives a sampled estimator of this expectation. More generally, an algorithm can construct advantage estimates from the full sampled group, in which case we directly define the empirical surrogate conditional on the current samples:
\[
\begin{aligned}
L_A(q)&=\frac1G\sum_{g=1}^G A^{(g)}\frac{q(\mathbf y^{(g)})}{p(\mathbf y^{(g)})},\\
\left.\nabla_\theta L_A(\pi_\theta(\cdot\mid x))\right|_{\theta=\theta_{\old}}
&=\frac1G\sum_{g=1}^G A^{(g)}\left.\nabla_\theta\log\pi_\theta(\mathbf y^{(g)}\mid x)\right|_{\theta=\theta_{\old}}.
\end{aligned}
\]
The key condition for extending the local surrogate is therefore that the algorithm's reward gradient at the starting point can be expressed in this rollout-level weighted form. Group-dependent advantage estimates are generally correlated with the current rollout, so the probability-ratio identity alone does not make this an unbiased estimator of the original expected-return gradient.

\paragraph{Sequence-level policy gradients.}
When a complete generated sequence is treated as one action, the REINFORCE sampled gradient corresponds to $A^{(g)}=R(x,\mathbf y^{(g)})-b(x)$ for a fixed prompt-level baseline $b(x)$. An actor--critic update that treats the sequence as one action and holds its advantage fixed while differentiating has the same form. This correspondence concerns the reward gradient and excludes gradients from additional regularizers.

\paragraph{Group-relative policy optimization.}
For outcome-supervised GRPO~\citep{shao2024deepseekmath} and GSPO~\citep{zheng2025gspo}, define the native group-relative advantage by
\[
A^{(g)}=\frac{R(x,\mathbf y^{(g)})-\bar R}{\sigma_R},\qquad
\bar R=\frac1G\sum_{g=1}^GR(x,\mathbf y^{(g)}),\qquad
\sigma_R=\left[\frac1G\sum_{g=1}^G(R(x,\mathbf y^{(g)})-\bar R)^2\right]^{1/2}.
\]
If $\sigma_R=0$, all advantages are defined as zero, so $\sum_gA^{(g)}=0$ always holds. Under outcome supervision, every token within a rollout shares $A^{(g)}$. Original GRPO averages the token loss within each rollout, whereas GSPO uses the length-normalized sequence probability ratio $[q(\mathbf y^{(g)})/p(\mathbf y^{(g)})]^{1/T_g}$. Ignoring the reference KL and differentiating at $\theta=\theta_{\old}$ with clipping inactive, both correspond to replacing $A^{(g)}$ in $L_A$ by $A^{(g)}/T_g$. If GRPO sums token losses rather than averaging within each rollout, it corresponds to $L_A$ without length normalization. These correspondences match parameter gradients only at the starting point and do not make the clipped objectives or their off-starting-point gradients identical.

\subsection{Proof of Proposition~\ref{prop:positive-dominance}}
\label{app:positive-only}

\paragraph{Independent-logit update and entropy derivative.}
Restrict the complete-sequence space to be finite, associate each sequence with an independent parameter $z_i$, and let $p=\operatorname{softmax}(z)$. Fix a sampled group with zero-sum advantages $\sum_gA^{(g)}=0$ and perform ordinary gradient ascent on $L(z)=G^{-1}\sum_gA^{(g)}\log p(\mathbf y^{(g)})$ by $z(\alpha)=z+\alpha\nabla_zL(z)$, using the same learning rate $\alpha$ for every coordinate. The gradient with respect to each logit and the corresponding entropy derivative are
\begin{equation}
\begin{aligned}
\frac{\partial L}{\partial z_i}&=\frac1G\sum_gA^{(g)}\bigl(\mathbf 1\{\mathbf y_i=\mathbf y^{(g)}\}-p_i\bigr)
=\frac1G\sum_gA^{(g)}\mathbf 1\{\mathbf y_i=\mathbf y^{(g)}\},\\
\frac{\partial H(p)}{\partial z_i}&=p_i(-\log p_i-H(p)).
\end{aligned}
\tag{A.9}\label{eq:logit-derivatives}
\end{equation}
Therefore
\begin{equation}
\left.\frac{dH(\operatorname{softmax}(z(\alpha)))}{d\alpha}\right|_{\alpha=0}=\sum_i\frac{\partial H(p)}{\partial z_i}\frac{\partial L}{\partial z_i}
=\sum_g\underbrace{\frac{e^{-s_g}A^{(g)}}G(s_g-H(p))}_{h_g^{\mathrm{SGD}}}.
\tag{A.10}\label{eq:sgd-entropy}
\end{equation}
Finite-dimensional smoothness also gives $H(\operatorname{softmax}(z(\alpha)))-H(p)=\alpha\sum_gh_g^{\mathrm{SGD}}+O(\alpha^2)$. The probability factor $e^{-s_g}$ arises from the entropy gradient with respect to logits, not from weighting a sampled rollout a second time. If $\sum_gA^{(g)}\ne0$, \eqref{eq:sgd-entropy} acquires the additional term $-\bigl(G^{-1}\sum_g A^{(g)}\bigr)\sum_ip_i^2(-\log p_i-H(p))$. The contribution $h_g^{\mathrm{SGD}}$ is therefore an additive attribution within a zero-sum whole-group update, not the complete entropy derivative obtained by training one rollout in isolation; it also cannot be reused automatically after deletion or length weighting.

\paragraph{Conditional dominance of a high-probability positive rollout.}
Assume $A_+>0>A_-$, $s_+<H(p)<s_-$, and equal entropy contributions $h_+=h_-<0$ in Proposition~\ref{prop:single-rollout}. By \eqref{eq:single-contribution}, this equality is equivalent to $A_+(H(p)-s_+)=|A_-|(s_--H(p))>0$. By \eqref{eq:sgd-entropy}, both rollouts decrease entropy, and their contribution magnitudes satisfy
\begin{equation}
\frac{|h_+^{\mathrm{SGD}}|}{|h_-^{\mathrm{SGD}}|}
=\frac{e^{-s_+}A_+(H(p)-s_+)}{e^{-s_-}|A_-|(s_--H(p))}
=e^{s_--s_+}>1.
\tag{A.11}\label{eq:appendix-positive-dominance}
\end{equation}
This proves \eqref{eq:positive-dominance}. The comparison concerns two rollouts whose entropy-decrease magnitudes before probability weighting are equal; it neither requires equal absolute advantages nor says that every positive-advantage rollout is more influential than every negative one.

\paragraph{Scale difference from the KL-proximal update.}
For the same sampled group with $\sum_gA^{(g)}=0$, \eqref{eq:aligned-score} satisfies $\tilde A_i=p_i^{-1}\partial L/\partial z_i$. \eqref{eq:closed-form-update} can therefore be written $q^{(\eta)}=\operatorname{softmax}(z+\eta\tilde A)$. The two logit-coordinate changes are
\begin{equation}
\Delta z_i^{\mathrm{KL}}=\frac{\eta}{p_i}\frac{\partial L}{\partial z_i},\qquad
\Delta z_i^{\mathrm{SGD}}=\alpha\frac{\partial L}{\partial z_i}.
\tag{A.12}\label{eq:update-scale-comparison}
\end{equation}
These are logit representatives of the corresponding distributions; adding the same constant to every coordinate does not change softmax. Section~\ref{sec:cov-deletion} also uses one global parameter $\eta$, but relative to the ordinary logit gradient its effective scale is $\eta/p_i$, whereas Section~\ref{sec:positive-only} uses the shared gradient multiplier $\alpha$. The inverse-probability factor in the former cancels the $p_i$ in the entropy gradient; the latter retains that probability factor. This is a difference between update rules, not a difference between having and not having a step size, and it does not make the independent-logit model more accurate than the KL-proximal model.

\paragraph{Scope of the conclusion.}
Proposition~\ref{prop:positive-dominance} is a first-order comparison under independent logits, zero-sum whole-group advantages, and the stated equal-magnitude condition. It is not a direct consequence of the KL-proximal update in Section~\ref{sec:cov-deletion} and does not establish positive-sample dominance for arbitrary finite steps or actual neural-network updates. It supplies additional motivation for the final positive-advantage filter. The maximum-cardinality property of \eqref{eq:cov-contribution-selection} applies only to the initial retained set with fixed original-group scores; the filtering in \eqref{eq:positive-filter} and recentering in \eqref{eq:selected-renorm} do not inherit its threshold constraint or any global optimality claim. These results are rigorous within their stated local models and guide method design, while the experiments in Sections~\ref{sec:main-results} and~\ref{sec:ablation} evaluate the complete method. They do not guarantee entropy preservation under shared-parameter AdamW updates.

\subsection{Proof of Proposition~\ref{prop:outside-paths}}
\label{app:outside-paths}

\paragraph{First-order change of zero-advantage sequences.}
With the native advantages fixed and satisfying $\sum_{g=1}^G A^{(g)}=0$, deleting a nonempty set $\mathcal D$ of positive-advantage rollouts leaves $\sum_{g\in\mathcal I_{\mathrm{keep}}}A^{(g)}=-\sum_{g\in\mathcal D}A^{(g)}<0$. Retention depends on rollout content and within-group comparisons, so the retained samples cannot generally be treated as fresh i.i.d. draws from $p$. These changes motivate reconsidering the advantage baseline after selection, but do not by themselves determine the probability weights.

Fix a prompt $x$, a realized group of rollouts sampled i.i.d. from $p$, a nonempty retained set $\mathcal I_{\mathrm{keep}}$, and a bounded post-deletion advantage $\tilde A_i^{\mathrm{keep}}$. To isolate normalization effects, assume the local update $q_{\mathrm{keep},i}^{(\eta)}=p_i\exp(\eta\tilde A_i^{\mathrm{keep}})/Z_{\mathrm{keep}}(\eta)$, where $Z_{\mathrm{keep}}(\eta)=\sum_jp_j\exp(\eta\tilde A_j^{\mathrm{keep}})$. This has the same KL-proximal form as Appendix~\ref{app:proofs}, but uses the accumulated retained advantages as its score rather than the sampling-aligned score in \eqref{eq:aligned-score}. The distribution-space update couples paths only through its normalizing denominator and contains no shared-parameter effects. Let $U=\{i:p_i>0,\ \tilde A_i^{\mathrm{keep}}=0\}$. The set $U$ contains sequences that receive no retained signal and may also include sampled sequences whose recentered advantage is zero or whose signals cancel; it must not be identified with the set of unsampled sequences. Under this update, $q_{\mathrm{keep},i}^{(\eta)}=p_i/Z_{\mathrm{keep}}(\eta)$ for $i\in U$. For $i\in U$, the unnormalized weight remains $p_i$, while normalization multiplies all such probabilities by the same factor $1/Z_{\mathrm{keep}}(\eta)$. Equivalently, every zero-advantage path receives the same additive log-probability offset $-\log Z_{\mathrm{keep}}(\eta)$, regardless of its similarity to the trained rollouts, and probability ratios between zero-advantage paths remain unchanged. In particular, any two paths $i,j\in U$ satisfy $q_{\mathrm{keep},i}^{(\eta)}/q_{\mathrm{keep},j}^{(\eta)}=p_i/p_j$: normalization cannot preferentially favor the one more similar to a rewarded rollout. The i.i.d. assumption concerns the original sampling group. The derivation is conditional on the realized group and retained set and does not require the retained samples to remain i.i.d.; excluding transfer through shared parameters is a separate assumption of the local update model. Using $Z_{\mathrm{keep}}(0)=1$ and $Z_{\mathrm{keep}}'(0)=\mathbb E_p[\tilde A^{\mathrm{keep}}]$ gives
\begin{equation}
\left.\frac{dq_{\mathrm{keep},i}^{(\eta)}}{d\eta}\right|_0=-p_i\mathbb E_p[\tilde A^{\mathrm{keep}}],\qquad
\left.\frac{d\log q_{\mathrm{keep},i}^{(\eta)}}{d\eta}\right|_0=-\mathbb E_p[\tilde A^{\mathrm{keep}}],\quad i\in U.
\tag{A.13}\label{eq:zero-score-change}
\end{equation}
This proves \eqref{eq:outside-prob-effect}. At $\eta=0$, the identical log-probability derivatives imply equal relative changes across zero-advantage paths, while their absolute changes are scaled by $p_i$. When $U$ is nonempty, every sequence in it has zero first-order probability change if and only if $\mathbb E_p[\tilde A^{\mathrm{keep}}]=0$. If $U$ is empty, there are no such sequences and one cannot infer a zero mean from a condition on an empty set. The result does not imply finite-step invariance: for a nontrivial post-deletion advantage with zero mean, typically $Z_{\mathrm{keep}}(\eta)>1$ for $\eta>0$, so zero-advantage sequences can still change at second and higher orders. Normalization here occurs only over the response space of a fixed prompt and does not prove cross-prompt or OOD invariance after a shared-network update.

\paragraph{Subset entropy and global entropy attribution.}
Let $H_U(q)=-\sum_{i\in U}q_i\log q_i$ be the unnormalized entropy of the subset $U$. \eqref{eq:zero-score-change} yields
\begin{equation}
\left.\frac{dH_U(q_{\mathrm{keep}}^{(\eta)})}{d\eta}\right|_0
=\mathbb E_p[\tilde A^{\mathrm{keep}}]\bigl(\sum_{i\in U}p_i-H_U(p)\bigr),\qquad
-\sum_{i\in U}q_{\mathrm{keep},i}'(0)\log p_i=-\mathbb E_p[\tilde A^{\mathrm{keep}}]H_U(p).
\tag{A.14}\label{eq:subset-entropy}
\end{equation}
The left expression is the derivative of the unnormalized subset entropy itself. The right expression is the contribution assigned to $U$ after the constant term has been canceled globally using $\sum_iq_{\mathrm{keep},i}'(0)=0$. They must not be conflated, and the sign of $\mathbb E_p[\tilde A^{\mathrm{keep}}]$ alone does not determine the direction of global entropy. The general global expression remains $-\Cov_p(\tilde A^{\mathrm{keep}},\log p)$ and reduces to the empirical covariance in \eqref{eq:covariance-conclusion} only for the original zero-sum score specified in Appendix~\ref{app:proofs}.

\paragraph{Which zero-mean condition probability-weighted recentering satisfies.}
For the nonempty retained set in \eqref{eq:selected-renorm}, replacing the native advantages in the accumulated post-deletion advantage by the recentered advantages gives
\begin{equation}
\tilde A_i^{\mathrm{keep}}
=\sum_{g\in\mathcal I_{\mathrm{keep}}}\bar A_{\mathrm{keep}}^{(g)}\mathbf 1\{\mathbf y_i=\mathbf y^{(g)}\},\qquad
\mathbb E_p[\tilde A^{\mathrm{keep}}]
=\sum_{g\in\mathcal I_{\mathrm{keep}}}e^{-s_g}\bar A_{\mathrm{keep}}^{(g)}=0.
\tag{A.15}\label{eq:probability-weighted-zero}
\end{equation}
The last equality follows directly from $\bar A_{\mathrm{keep}}^{(g)}=A^{(g)}-b_{\mathrm{keep}}$ and $b_{\mathrm{keep}}=\sum_{g\in\mathcal I_{\mathrm{keep}}}e^{-s_g}A^{(g)}/\sum_{g\in\mathcal I_{\mathrm{keep}}}e^{-s_g}$. \eqref{eq:selected-renorm} therefore eliminates the first-order normalization drift of zero-advantage sequences in the KL-proximal model driven by this particular post-deletion advantage. The probability weights arise from the old-policy expectation in \eqref{eq:zero-score-change}: \eqref{eq:selected-renorm} imposes this weighted zero-mean condition on the retained score without assuming that it held before deletion. Subtracting a common baseline preserves pairwise differences between retained advantages; the weights calibrate the retained score rather than reconstruct the original sampling distribution through importance weighting. Advantage contributions are accumulated by sampled index; when sequences repeat, the denominator $\sum_{g\in\mathcal I_{\mathrm{keep}}}e^{-s_g}$ counts each sampled occurrence and need not equal the probability mass of the distinct retained sequences.

\section{Experimental Details}
\label{app:experimental-details}

\newtcolorbox{diversitypromptbox}{%
  breakable,
  colback=orange!4,
  colframe=orange!70!black,
  boxrule=0.75pt,
  leftrule=2.2pt,
  arc=1.5pt,
  left=5pt,
  right=5pt,
  top=4pt,
  bottom=4pt,
  fonttitle=\bfseries\small,
  coltitle=orange!70!black,
  colbacktitle=orange!4,
  title={\textsc{GLM-5.2 summarizer prompt}}
}

\subsection{Experimental Setup}
\label{app:experimental-setup}

\paragraph{Training settings.}
We conduct experiments on Qwen3-1.7B, Qwen3-4B, and Qwen3-8B~\citep{qwen3technicalreport}, using DAPO-17K mathematics data~\citep{yu2025dapo} for reinforcement learning. All model sizes use the same training configuration and are trained for one epoch. Each training batch contains 128 prompts with eight rollouts per prompt, totaling 1,024 candidate rollouts before selection. The maximum prompt and response lengths are 2,048 and 8,192 tokens, respectively, and overlong prompts are filtered. Rollouts are sampled with vLLM at temperature 1.0 and top-$p$ 1.0. Original GRPO uses all eight rollouts; GRPODropout removes selected rollouts and recenters the retained advantages before updating the policy. We use AdamW with learning rate $10^{-6}$ and weight decay 0.01, a clipping width of 0.2, and a policy loss averaged over valid response tokens. The original GRPO baseline uses no additional entropy bonus or reference-policy KL regularization.

\paragraph{Baselines.}
We use original GRPO~\citep{shao2024deepseekmath} as the base control and compare representative methods from the two categories discussed in Related Work. The first category selects or masks token-level training signals: Beyond8020~\citep{wang2025beyond8020} optimizes only the approximately 20\% highest-entropy tokens, while Clip-Cov~\citep{cui2025entropymechanism} masks updates for a subset of high-covariance tokens. The second category modifies the RL objective or update rule: KL-Cov~\citep{cui2025entropymechanism} applies KL constraints to selected high-covariance tokens, while DAPO~\citep{yu2025dapo} uses Clip-Higher to relax the upper clipping bound and facilitate exploration of low-probability actions. These baselines cover token-level data usage, regularization, and clipping, allowing us to compare complete-rollout selection with different strategies for mitigating entropy collapse. GRPODropout operates on data usage, applying the original RL loss to retained rollouts with recentered advantages while preserving its objective form.

\paragraph{Evaluation settings.}
We evaluate on ten benchmarks: GSM8K~\citep{cobbe2021verifiers}, MATH-500~\citep{lightman2023verify}, OlympiadBench~\citep{he2024olympiadbench}, AMC23, AIME24, AIME25, and HMMT25 for mathematical reasoning, and ARC-C~\citep{clark2018arc}, MMLU-Pro~\citep{wang2024mmlupro}, and SuperGPQA~\citep{supergpqa2025dataset} for general question answering. The latter evaluates generalization beyond the mathematics training domain. All evaluations use a maximum generation length of 8,192 tokens, temperature 0.6, top-$p$ 0.95, top-$k$ 20, and a fixed random seed. GSM8K, MATH-500, OlympiadBench, and the three QA benchmarks use one sampled response per question and report Mean@1. For the smaller AMC23, AIME24, AIME25, and HMMT25 test sets, we sample 32 responses per question through repeated single-response requests and report Mean@32, averaging correctness over the 32 responses and then over questions. Mathematical prompts request step-by-step reasoning and a final answer in \texttt{\textbackslash boxed\{\}}. QA prompts request a final option letter, which is extracted, case-normalized, and compared with the reference label. Avg. is the unweighted mean accuracy across all ten benchmarks.

\subsection{Diversity-analysis protocol}
\label{app:table3-diversity}

This subsection records the protocol used to compute the path-diversity columns in Table~\ref{tab:precision-exploration}. The analysis covers 130 unique questions in AMC23 (40), AIME24 (30), AIME25 (30), and HMMT25 (30), with 32 sampled responses per question and system. Pass@32 is computed per question within each benchmark and then averaged equally across AMC23, AIME24, AIME25, and HMMT25. The Entropy column uses token-mean actor entropy over valid response tokens at the final checkpoint, averaged equally across the four benchmarks. All path-diversity aggregates are unweighted means over questions.

\paragraph{Question partition and answer dispersion.}
Let $c_{m,q}$ be the number of correct responses among the 32 rollouts for method $m$ on question $q$. We define the common high-correctness set as
\[
  \mathcal Q_{\geq 8}=\bigcap_m\{q:c_{m,q}\geq 8\},
  \qquad |\mathcal Q_{\geq 8}|=65,
\]
and the common low-correctness set as
\[
  \mathcal Q_{<8}=\bigcap_m\{q:c_{m,q}<8\},
  \qquad |\mathcal Q_{<8}|=37.
\]
Thus the threshold is based on the original rollout correctness counts, and the same question sets are used for every method. The remaining 28 questions have mixed membership across systems and are not used in the two split columns. For each method and question, we compute
\[
  W_{m,q}=\left|\left\{\operatorname{strip}(a_i):r_i=0,\ \operatorname{strip}(a_i)\neq\emptyset\right\}\right|,
\]
where $a_i$ is the extracted final answer and $r_i$ is its correctness indicator. $W$ measures answer-string dispersion among incorrect final answers and can reveal recurring answer patterns.

\paragraph{Summary generation and embedding.}
For each rollout, the summarizer receives the original question and the full generated response after removing explicit \texttt{<think>} tags. We use the model \texttt{GLM-5.2}, temperature $0$, with thinking disabled. The summarizer is instructed to retain only the reasoning path that actually leads to the final boxed answer, place abandoned ideas in a separate list, and return strict JSON. Its exact prompt is:

\begin{diversitypromptbox}
\begin{Verbatim}[fontsize=\scriptsize,breaklines=true,breakanywhere=true]
You compress ONE chain-of-thought solution to a math problem.

The trace is a stream-of-consciousness: the solver often floats several ideas,
tries some, abandons most, and finally carries ONE line of reasoning through to
the boxed answer. Your job is to separate the two:

- FINAL PATH: the reasoning that was ACTUALLY carried through to produce the
  final (boxed) answer. This is the solution's real solving path. Identify it by
  looking at how the last, answer-producing derivation actually went -- NOT by
  what was merely proposed earlier.
- ATTEMPTED: ideas that were floated, tried, then dropped / not used for the
  final answer. Record each briefly with why it was abandoned.

Be faithful. Do NOT invent steps. If the solver kept switching, the FINAL PATH
is the one whose result equals the boxed answer at the very end.

Output STRICT JSON only:
{
  "final_path": {
    "summary": "<3-5 sentences: the actual path that yielded the answer>",
    "method": "<short tag: main machinery, e.g. cartesian-algebra / polar-form / calculus / geometric-series-identity>",
    "angle": "<short tag: the entry point / first decisive framing that led to the answer>",
    "key_trick": "<short tag: the crucial simplification/insight used, or 'none'>"
  },
  "attempted": [
    {"idea": "<short: an approach that was tried>", "why_abandoned": "<short>"}
  ]
}
\end{Verbatim}
\end{diversitypromptbox}

The returned summary is encoded with the local \texttt{bge-m3} checkpoint with L2 normalizations, and float32 arithmetic. For each question, Summary distance is the mean pairwise cosine distance over all 32 summary embeddings. Vendi is computed from the eigenvalues of the symmetrized Gram matrix divided by 32 as $\exp(-\sum_j\lambda_j\log\lambda_j)$. Both measures therefore use all 32 summaries, including correct and incorrect responses.

\end{document}